\documentclass[9pt]{article}
\usepackage{amsmath,graphicx,mlspconf}

\usepackage{cite}
\usepackage{amsmath,amssymb,amsfonts}
\usepackage{amsthm}

\usepackage{algorithmic}
\usepackage{subcaption}
\usepackage{graphicx}
\usepackage{textcomp}
\usepackage{xcolor}
\usepackage{url}
\usepackage{mathtools}
\usepackage{enumitem}
\usepackage{booktabs}
\usepackage{multirow}
\usepackage{algorithm}
\usepackage{algorithmic}

\usepackage[algo2e,linesnumbered,lined, ruled]{algorithm2e}

\usepackage{tikz}
\definecolor{myblue}{RGB}{48 107 196}

\newtheorem{theorem}{Theorem}[section]

\newtheorem{definition}[theorem]{Definition}
\newtheorem{remark}{Remark}

\newtheorem{manualtheoreminner}{Theorem}
\newenvironment{manualtheorem}[1]{%
  \renewcommand\themanualtheoreminner{#1}%
  \manualtheoreminner
}{\endmanualtheoreminner}

\providecommand{\By}{{\mathbf{y}}}
\providecommand{\Bx}{{\mathbf{x}}}

\providecommand{\Bd}{{\mathbf{d}}}
\providecommand{\drift}{{\mathbf{f}}}

\providecommand{\bA}{{\mathbf{A}}}
\providecommand{\Bxt}{{\mathbf{x}_t}}

\DeclareMathOperator*{\argmin}{arg\,min}

\providecommand{\tmin}{{t_\text{min}}}
\providecommand{\tmax}{{t_\text{max}}}
\providecommand{\N}{\mathbb{N}}
\providecommand{\R}{\mathbb{R}}
\providecommand{\E}{\mathbb{E}}
\providecommand{\X}{\mathcal{X}}
\providecommand{\Y}{\mathcal{Y}}

\toappear{\textbf{Preprint submitted to IEEE MLSP 2024}}

\title{Convergence Properties of Score-Based Models for Linear Inverse Problems Using Graduated Optimisation}
\name{
    Pascal Fernsel$^{\star}$%
    \qquad \v{Z}eljko Kereta$^{\dagger}$%
    \qquad Alexander Denker$^{\dagger}$\thanks{\v{Z}. K. is supported by the UK EPSRC grant EP/X010740/1. A.D. is funded by EPSRC programme grant EP/V026259/1. P. F. acknowledges funding by the Deutsche Forschungsgemeinschaft (DFG) – Project number 281474342. Corresponding author: \protect\url{p.fernsel@uni-bremen.de}}%
}
\address{%
    $^{\star}$\textit{Center for Industrial Mathematics},
\textit{University of Bremen}  \\%
    $^{\dagger}$\textit{Department of Computer Science}, \textit{University College London}
}

\begin{document}

\maketitle

\begin{abstract}
The incorporation of generative models as regularisers within variational formulations for inverse problems has proven effective across numerous image reconstruction tasks. However, the resulting optimisation problem is often non-convex and challenging to solve. In this work, we show that score-based generative models (SGMs) can be used in a graduated optimisation framework to solve inverse problems. We show that the resulting graduated non-convexity flow converge to stationary points of the original problem and provide a numerical convergence analysis of a 2D toy example. We further provide experiments on computed tomography image reconstruction, where we show that this framework is able to recover high-quality images, independent of the initial value. The experiments highlight the potential of using SGMs in graduated optimisation frameworks. The code is available\footnote{\url{https://github.com/alexdenker/GradOpt-SGM}}.
\end{abstract} 
\section{Introduction}
Many problems in image reconstruction can be formulated as linear inverse problems, where the goal is to recover an image $\Bx \in \X$ given noisy measurements $\By^\delta$ that are related via \vspace{-1ex}
\begin{align}
    \By^\delta = \bA\Bx + \epsilon.
\end{align}
Here $\bA:\X \to \Y$ is a linear forward operator and $\epsilon \in \Y$ the noise. Inverse problems are often ill-posed and require regularisation to stabilise the reconstruction. 
Variational regularisation is a popular framework to address ill-posedness \cite{scherzer2009variational}. It formulates the reconstruction as an optimisation problem \vspace{-1ex}
\begin{align}\label{eqn:varreg}
    \hat{\Bx} \in \argmin_\Bx \{\mathcal{D}(\bA\Bx,\By^\delta) + \alpha \mathcal{R}(\Bx)\},
\end{align}\\[-3ex]
where $\mathcal{D}:\Y \times \Y \to \R_{\ge 0}$ quantifies fitness to the measurements, $\mathcal{R}: \X \to \R_{\ge 0}$ is a regulariser, and $\alpha\geq0$ balances the two terms. 
For additive Gaussian noise the datafit is typically chosen as $\mathcal{D}(\bA \Bx, \By^\delta) = \frac{1}{2} \| \bA \Bx - \By^\delta \|_2^2$. Finding a suitable regulariser is a difficult task. 
In recent years a wide variety of deep learning methods have been developed that aim to learn a regulariser from a given dataset, see  \cite{arridge2019solving} for a review.

The statistical perspective on inverse problems identifies the regulariser with the negative log-likelihood of a prior distribution $\pi(\Bx)$, i.e. $\mathcal{R}(\Bx) = -\log \pi(\Bx)$. 
In this context, learning a prior can be formulated as learning a parametrised distribution $p_\theta(\Bx)$ as a proxy for $\pi(\Bx)$ \cite{duff2024regularising}.
If we have access to the likelihood, we can consider the optimisation problem \vspace{-1ex}
\begin{align}
    \label{eq:generative_reg}
    \hat{\Bx} \in \argmin_\Bx \{\mathcal{D}(\bA\Bx, \By^\delta) - \log p_\theta(\Bx)\},
\end{align}\\[-2.5ex]
as an instance of variational regularisation. 
Generative regularisers of this form have been extensively studied for various inverse problems \cite{duff2024regularising}. However, as the negative log-likelihood of the generative model is often highly non-convex, \eqref{eq:generative_reg} is a challenging optimisation problem with many local minima and it strongly depends on the initialisation.


We propose to tackle this problem using score-based generative models (SGMs) \cite{song2021scorebased} as regularisers. SGMs learn a sequence of gradually perturbed distributions, starting at the data distribution and terminating in pure noise. It was recently observed that this sequence can be used in annealed Langevin sampling \cite{song2019generative,sun2023provable} and graduated optimisation \cite{kobler2023learning}. Graduated optimisation \cite{blake1987visual}, also known as continuation methods \cite{mobahi2015link}, is a heuristic for dealing with non-convex problems in which the objective function is replaced with a convex surrogate which can be solved efficiently. The surrogate objective is then gradually transformed to the original non-convex objective.

In this work, we exploit the connection between SGMs and graduated optimisation to solve the underlying non-convex optimisation problem and avoid local minima. 
We showcase our algorithms on computed tomography.

\section{Background}
\subsection{Graduated Optimisation}
\label{sec:graduated_optimisation}

Graduated non-convexity is a heuristic global optimisation method for solving non-convex minimisation problems, which creates a sequence of surrogate optimisation problems that approximate the original problem through gradual levels of smoothing or convexification \cite{blake1987visual}.
Namely, for a non-convex function $f: \X \to \R_{\ge 0}$ let $F: \X \times [\tmin, \tmax] \to \R_{\ge 0}$, such that $F(\Bx, \tmin) = f(\Bx)$ and that $F(\Bx, \tmax)$ is convex with a unique minimiser. 
The optimisation protocol consists of $I\in\N$ iterations over $\tmax = t_1 > \dots > t_I = \tmin$. 
In step $i$ we find the minimiser $\Bx_i$ of $F(\Bx, t_i)$, using $\Bx_{i-1}$ as the starting value. 
The process is terminated by minimising the original function $f(\Bx)$ using $\Bx_{I-1}$ as initialisation. 

The success of graduated optimisation strongly depends on the construction of the embedding function $F(\Bx, t)$. Under specific constraints on the type of embedding and the class of non-convex function under consideration, it is possible to get global convergence results \cite{hazan2016graduated}.



\begin{algorithm}[t]
\setcounter{ALC@unique}{0}
\caption{\texttt{Graduated non-convexity flow with step size rule}}
\label{alg:gradientLikeAlgorithm_v3}
\begin{algorithmic}[1] 
\STATE{\textbf{Initialise:} $\!\displaystyle \Bx_1\!\! \in\!\! \R^n, \tmax \!\!=\! t_1\!\! >\!\! \dots \!\!>\! t_I \!= \!\tmin, I \!\!\in\!\! \N, c\!\! \in\!\! (\!0,\!1\!)$}
    \FOR{$i=1, \dots, I-1$}
    \STATE{$\Bd_i = - t_i \nabla_\Bx F(\Bx_i, t_i)$}
    \STATE{\textbf{Find:} $\displaystyle \lambda_i$ s.t.$f(\Bx_i+\!\lambda_i \Bd_i)\! \leq \!f(\Bx_i)\!+\!c \lambda_i \langle \nabla_\Bx f(\Bx_i), \Bd_i\rangle_{\R^n}$}
    \STATE{ $\displaystyle \Bx_{i+1} = \Bx_i + \lambda_i \Bd_i$}
    \ENDFOR
    \STATE{\textbf{Output:} $\displaystyle \Bx_I$}
\end{algorithmic}
\end{algorithm}

\begin{algorithm}[t]
\setcounter{ALC@unique}{0}
\caption{\texttt{Gradient-like Method}}
\label{alg:gradientLikeAlgorithm}
\begin{algorithmic}[1] 
\STATE\label{alg:gradientLikeAlgorithm_Line1}{\textbf{Initialize:} $\!\displaystyle \ \Bx_1 \!\in\! \R^n, I\!\in\! \N, c\!\in\! (0,\!1\!), \beta\!\in\! (0,\!1\!), \varepsilon\!>\!0,\ i\!=\!1$}
    \WHILE{$1\leq i \leq I \land \Vert \nabla_\Bx f(\Bx_i) \Vert > \varepsilon$}
    \STATE\label{alg:gradientLikeAlgorithm_innerProduct}{\textbf{Determine:} $\displaystyle \Bd_i \in \R^n \quad \text{s.t.} \quad \langle \nabla_\Bx f(\Bx_i), \Bd_i\rangle_{\R^n} <0$}
    \STATE\label{alg:gradientLikeAlgorithm_Armijo}{\textbf{Determine:} $\displaystyle \lambda_i \coloneqq \max\{\beta^\ell  |  \ell=0,1,2,3,\dots \}$ s.t. $ f(\Bx_i + \lambda_i \Bd_i) \leq f(\Bx_i) + c \lambda_i \langle \nabla_\Bx f(\Bx_i), \Bd_i\rangle_{\R^n}$}
    \STATE{$\Bx_{i+1} \gets \Bx_i + \lambda_i \Bd_i$}
        \STATE{$i \gets i+1$}
    \ENDWHILE
\end{algorithmic}
\end{algorithm}

\subsection{Score-based Generative Models}
\label{sec:score_based_gen}
Score-based generative models (SGMs) have emerged as a powerful tool for generative modelling \cite{song2021scorebased,dhariwal2021diffusion}.  
SGMs consist of a (predefined) forward process, during which noise is gradually added; and a learnable reverse process, which allows transforming noise into samples. 
The forward process is prescribed by an It\^{o} stochastic differential equation (SDE)\vspace{-1ex}
\begin{align}
    d \Bxt = \drift(\Bxt, t) dt + g(t) d \mathbf{w}_t, \quad \Bx_0 \sim p_0 := \pi.
\end{align}
The drift function $\drift:\X \times \R \to \X$ and the diffusion function $g:\R \to \R$ are chosen such that the terminal distribution approximates a Gaussian, i.e. $p_T \approx \mathcal{N}(0,\mathbf{I})$. Under certain conditions there exists a reverse diffusion process\vspace{-1ex}
\begin{align}
    d \Bx_t = \left[\drift(\Bx_t, t) - g(t)^2 \nabla_{\Bx_t} \log p_t(\Bx_t) \right] dt + g(t) d \mathbf{w}_t,
\end{align}
which runs backwards in time~\cite{anderson1982reverse}. 
SGMs approximate the score function $\nabla_{\Bx_t} \log p_t(\Bx_t)$ by a neural network $s_\theta(\Bx_t, t)$, which can be trained using denoising score matching \cite{vincent2011connection}\vspace{-1ex}
\begin{align*}
     \min_\theta \mathop{\E}_{\substack{t \sim U(0,1), \,  \Bx_0 \sim \pi \\ \Bx_t \sim p_{t|0}(\Bx_t|\Bx_0)}} \left[ \| s_\theta(\Bx_t, t) - \nabla_{\Bx_t} \log p_{t|0}(\Bx_t|\Bx_0) \|_2^2  \right].
\end{align*}
For an affine drift $\drift$, the transition kernel $p_{t|0}$ is given as $p_{t|0}(\Bx_t|\Bx_0)= \mathcal{N}(\Bx_t; \gamma_t \Bx_0, \nu_t^2 \mathbf{I})$ with parameters $\gamma_t, \nu_t >0$ \cite{sarkka2019applied}. 
The SGM learns all intermediate distributions, i.e.\vspace{-1ex}
\begin{align}
    \label{eq:noisy_prior}
    p_\theta(\Bx_t, t) \approx p_t(\Bx_t) = \int \pi(\Bx_0) p_{t|0}(\Bx_t | \Bx_0) d \Bx_0. 
\end{align}
Intermediate distributions are similar to Gaussian homotopy \cite{mobahi2015link}, as SDE without a drift gives $p_t(\cdot) = (\pi * \mathcal{N}(0, \nu_t^2 \mathbf{I}))(\cdot)$. 





\section{Methods} \label{sec:Methods}

\subsection{Graduated non-convexity Flow}
\label{sec:graduatednonconvexityflow}
Using the SGM framework we define negative log-likelihoods \vspace{-3.5ex}
\begin{align}
    \mathcal{R}(\Bxt, t) \coloneqq - \log p_t(\Bxt).
\end{align}\\[-3.5ex]
We expect that $\mathcal{R}(\cdot, \tmin)$ is highly non-convex so that the corresponding optimisation problem \vspace{-1ex}
\begin{align}
\label{eq:f}
\min_\Bx \left\{ f(\Bx) \coloneqq \mathcal{D}(\bA\Bx, \By^\delta) + \alpha_{\tmin} \mathcal{R}(\Bx, \tmin) \right\},
\end{align}\\[-3ex]
is challenging to solve.
However, due to the convolution with a Gaussian in \eqref{eq:noisy_prior}, there exists a $\Tilde{t} < \infty$ such that $\mathcal{R}(\cdot, \Tilde{t})$ is convex for all $t>\Tilde{t}$ \cite{kobler2023learning}. Since $\mathcal{D}(\bA \Bx, \By^\delta)$ is convex, the resulting optimisation problem is also convex at $\tmax$ and can be solved using common convex optimisation methods. This motivates using SGMs as an embedding in a graduate optimisation scheme, giving
a sequence of optimisation problems\vspace{-1ex}
\begin{align}
    \label{eq:seq_minimisation_1}
    \hspace{-4pt} \hat{\Bx}_{i+1}\!\in \argmin_\Bx \left\{ F(\Bx, t_i)\!\coloneqq\!\mathcal{D}(\bA\Bx, \By^\delta)\!+\!\alpha_{t_i} \mathcal{R}(\Bx, t_i) \right\},\!
\end{align}\\[-3ex]
where $\alpha_{t_i}>0$ is a regularisation parameter and the optimisation is initialised with $\hat{\Bx}_i$. Instead of exactly solving \eqref{eq:seq_minimisation_1} in each step, we perform one step of gradient descent,\vspace{-1ex} 
\begin{equation}
    \label{eq:graduated nonConvexity Flow}
    \begin{split}
    \Bx_{i+1} &= \Bx_i - \lambda t_i \nabla_\Bx F(\Bx_i, t_i) \\
    &= \Bx_i - \lambda t_i \bA^*(\bA\Bx_i - \By^\delta) + \lambda t_i \alpha_{t_i} s_\theta(\Bx_i, t_i),
        \end{split}
\end{equation}
where $s_\theta(\Bx_i, t_i) \approx \nabla_{\Bx_i} \log p_{t_i}(\Bx_i)$ is a pre-trained score-based model. 
This corresponds to Algorithm 2 in \cite{kobler2023learning} with a predefined smoothing schedule.

\subsection{Gradient-like Methods}\label{sec:Gradient-like Methods}

The critical issue in surrogate optimisation methods is ensuring the convergence to an optimum of the original problem.
For graduated non-convexity flow in Section \ref{sec:graduatednonconvexityflow} this boils down to update directions used in \eqref{eq:graduated nonConvexity Flow}, which are not ensured to point in the direction of steepest descent.
However, convergence of iterations of the form $\Bx_{i+1}=\Bx_i+\lambda_i\Bd_i$ can still be shown provided $\Bd_i$ is a descent direction, i.e. $\langle \nabla_\Bx f(\Bx_i), \Bd_i\rangle_{\R^n} <0$.
Iterative methods of this type are also known as \textit{gradient-like methods}  \cite[Section 8.3, p. 75]{geiger2013numerische} or sufficient descent methods \cite{an2011sufficient}. 

We consider the case when the maximal number of iterations $I$ (i.e. the fineness of the smoothing schedule) tends to infinity. 
Let $\{t_i\}_{i\in \N}\subset [\tmin, \tmax]$ be such that
\begin{align}\label{eq:t_i}
    t_1 = \tmax, && t_{i+1} \leq t_i \quad \forall i\in \N, && \displaystyle \lim_{i\to \infty}t_i = \tmin.
\end{align}\\[-3ex]
To show the convergence of $\{\Bx_i\}_{i\in\N}$ we use the following notion of gradient-like directions.
\begin{definition}[Gradient-like Directions \cite{geiger2013numerische}]\label{def:GradientLikeDirection}
    Let $f: \R^n \to \R$ be continuously differentiable and $\{\Bx_i\}_{i\in \N}\subseteq \R^n.$ A sequence $\{\Bd_i\}_{i\in \N}\subseteq \R^n$ is called \textit{gradient-like} with respect to $f$ and $\{\Bx_i\}_{i\in \N},$ if for every subsequence $\{\Bx_{i_j}\}_{j \in \N},$ converging to a non-stationary point of $f,$ there exist $\varepsilon_1,\varepsilon_2>0$ and $N^*\in \N$  such that
    \begin{enumerate}[label=(\alph*)]
        \item\label{def:GradientLikeDirection:a} $\Vert \Bd_{i_j} \Vert\leq \varepsilon_1, \ \forall j\in \N$ and 
        \item\label{def:GradientLikeDirection:b} $\langle \nabla_\Bx f(\Bx_{i_j}), \Bd_{i_j}\rangle_{\R^n} \leq -\varepsilon_2, \ \text{for}\ j\geq N^*.$
    \end{enumerate}
\end{definition}

Under these conditions every limit point of iterates produced by a gradient-like algorithm is a stationary point of $f$ (see \cite[Theorem 8.9]{geiger2013numerische}).
In order to show convergence we also require $\nabla_\Bx \mathcal{R}(\Bx, \cdot)$ to be Lipschitz continuous, which ensures that search directions $\{\Bd_i\}_{i\in \N}$ defined in \eqref{eq:d_i} are gradient-like with respect to $f$ in \eqref{eq:f} and iterates $\{\Bx_i\}_{i\in \N}.$ With this, we have the following convergence result.

\begin{figure*}[t]
    \centering
    \begin{subfigure}{.31\linewidth}
        \centering
        \includegraphics[width = \linewidth]{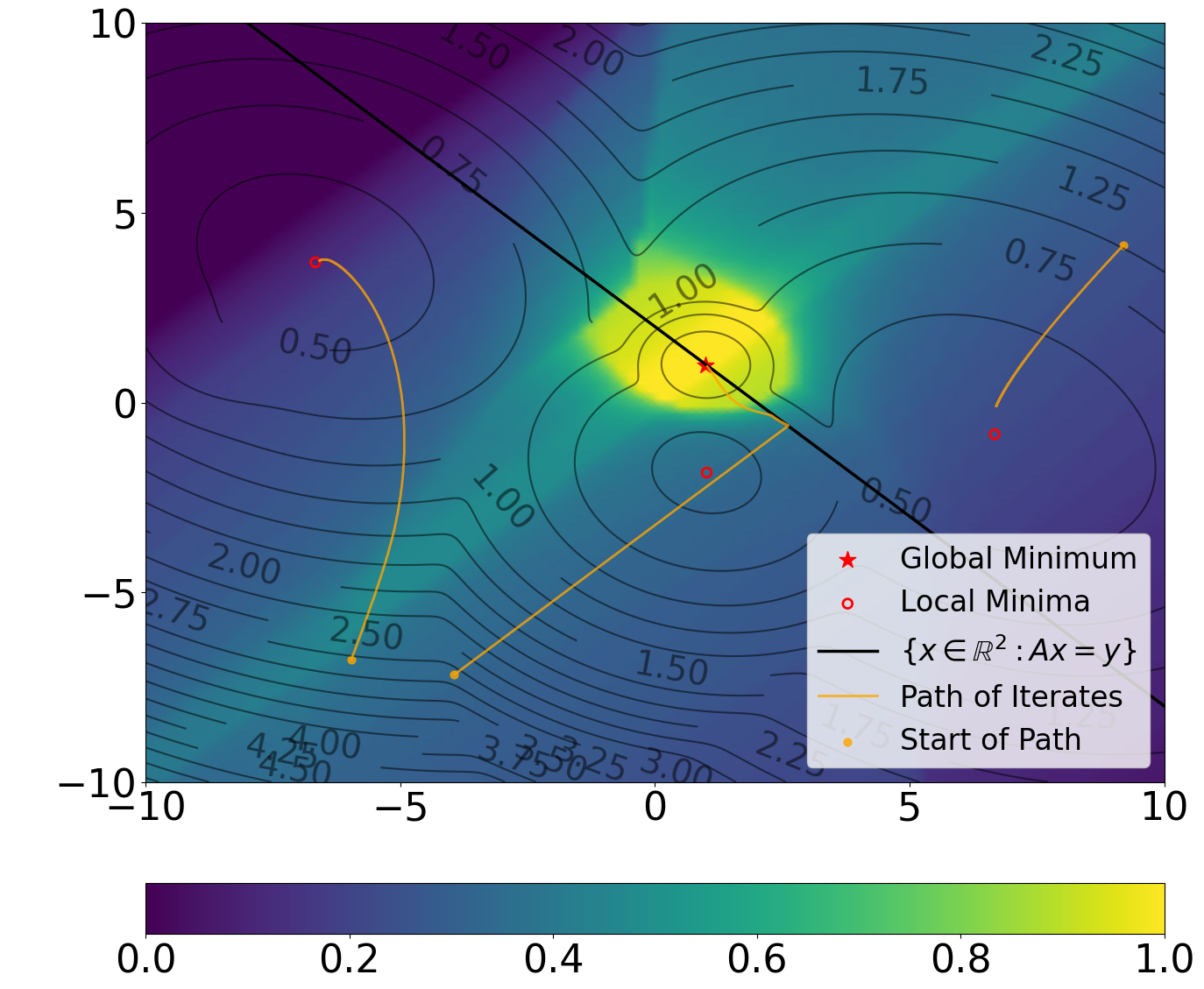}
        \caption{}
        \label{fig:2DIP:KoblerPock:xyplane}
    \end{subfigure}
    \begin{subfigure}{.31\linewidth}
        \centering
        \includegraphics[width = \linewidth]{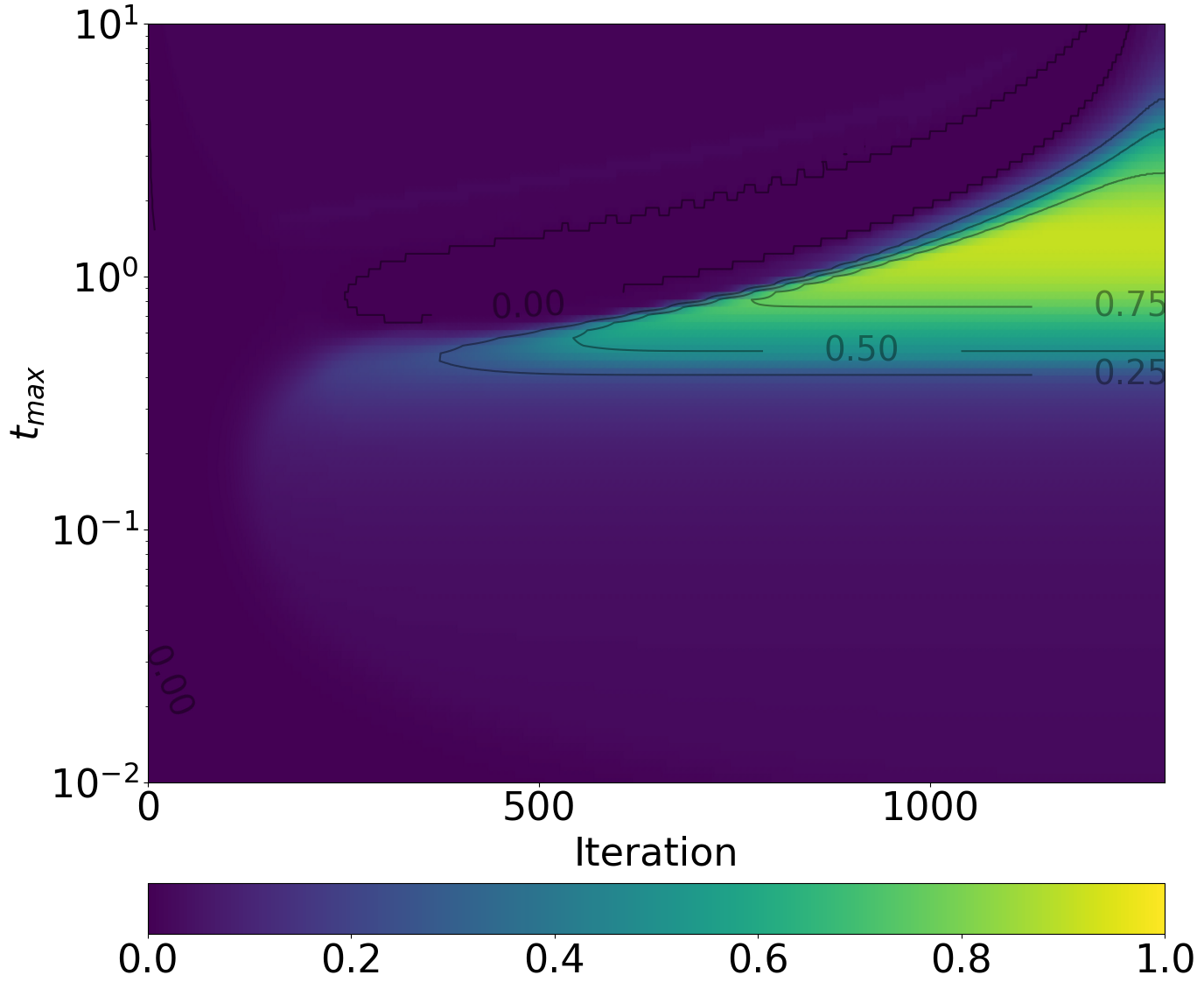}
        \caption{}
        \label{fig:2DIP:KoblerPock:globalMin}
    \end{subfigure}
    \begin{subfigure}{.31\linewidth}
        \centering
        \includegraphics[width = \linewidth]{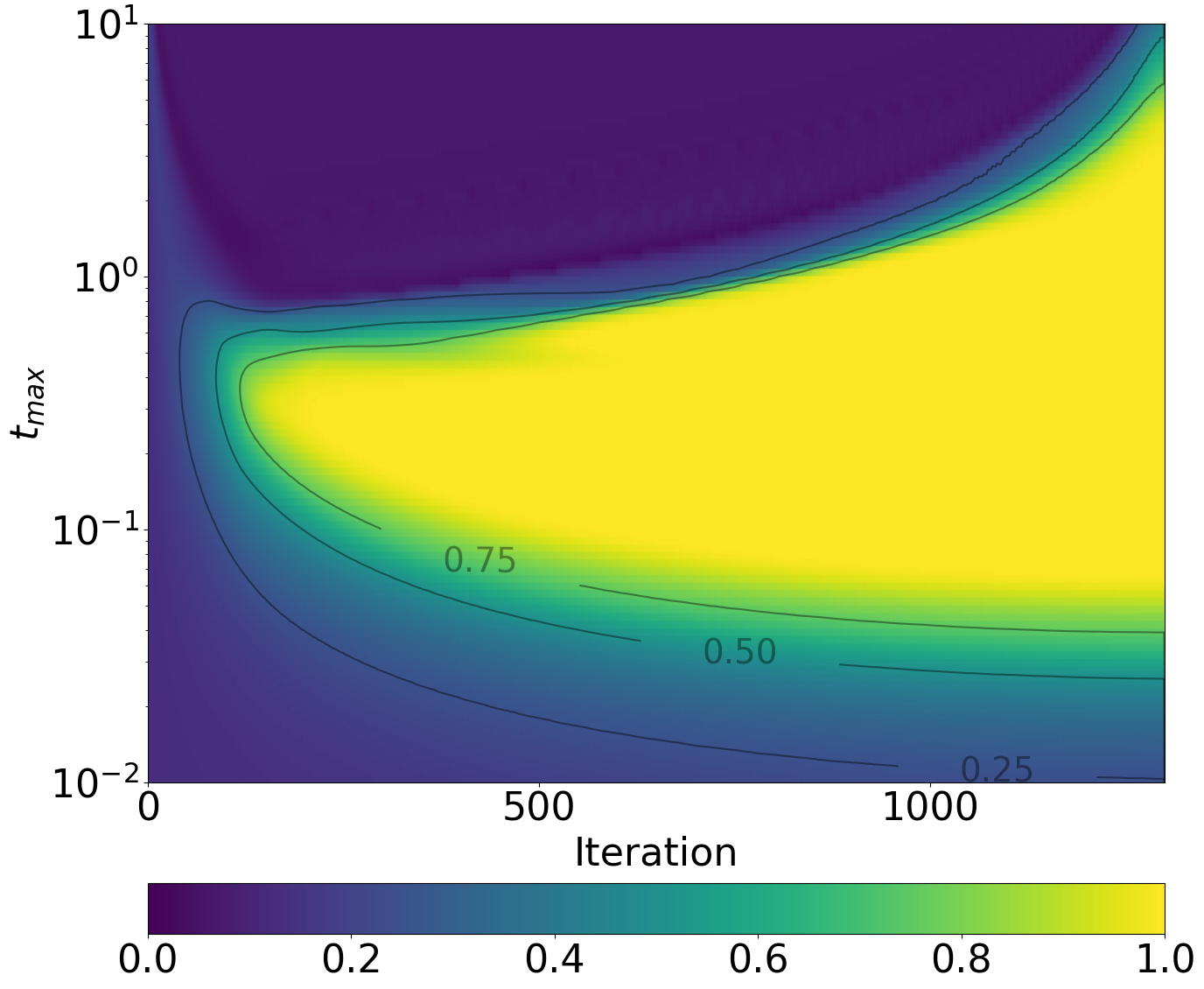}
        \caption{}
        \label{fig:2DIP:KoblerPock:localMin}
    \end{subfigure}
    \hfill
    \begin{subfigure}{.313\linewidth}
        \centering
        \includegraphics[width = \linewidth]{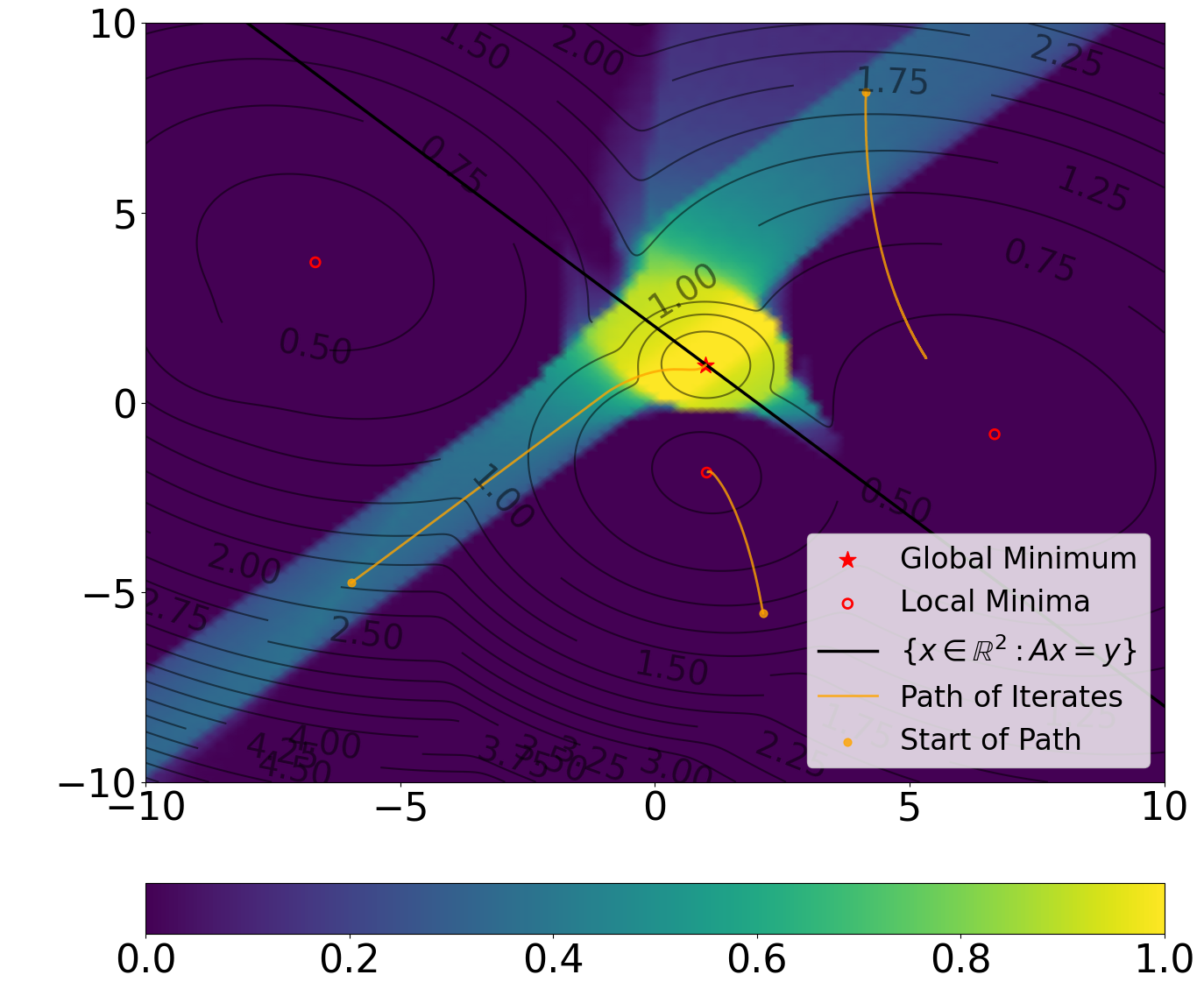}
        \caption{}
        \label{fig:2DIP:gradientLike:xyplane}
    \end{subfigure}
    \begin{subfigure}{.317\linewidth}
        \centering
        \includegraphics[width = \linewidth]{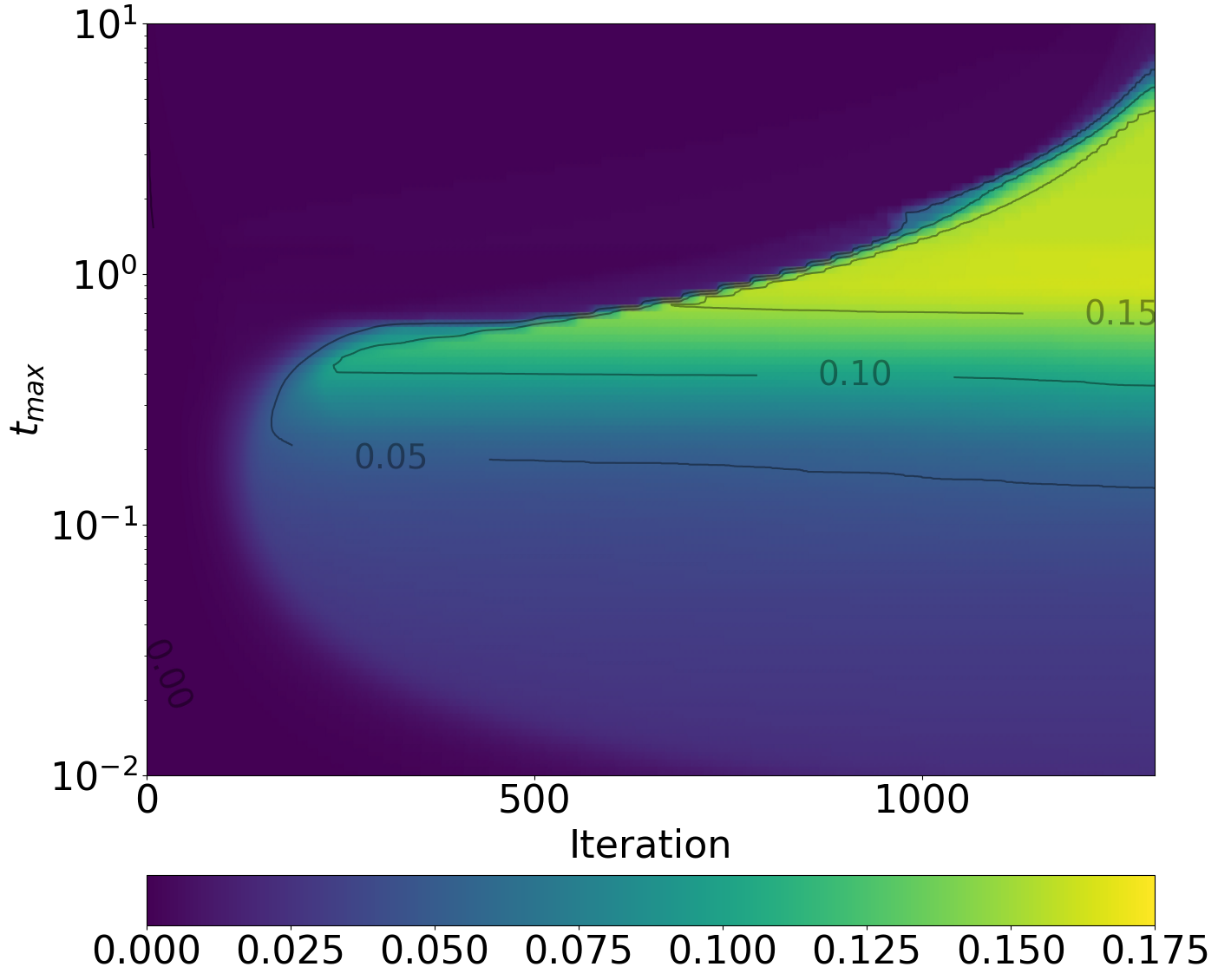}
        \caption{}
        \label{fig:2DIP:gradientLike:globalMin}
    \end{subfigure}
    \begin{subfigure}{.31\linewidth}
        \centering
        \includegraphics[width = \linewidth]{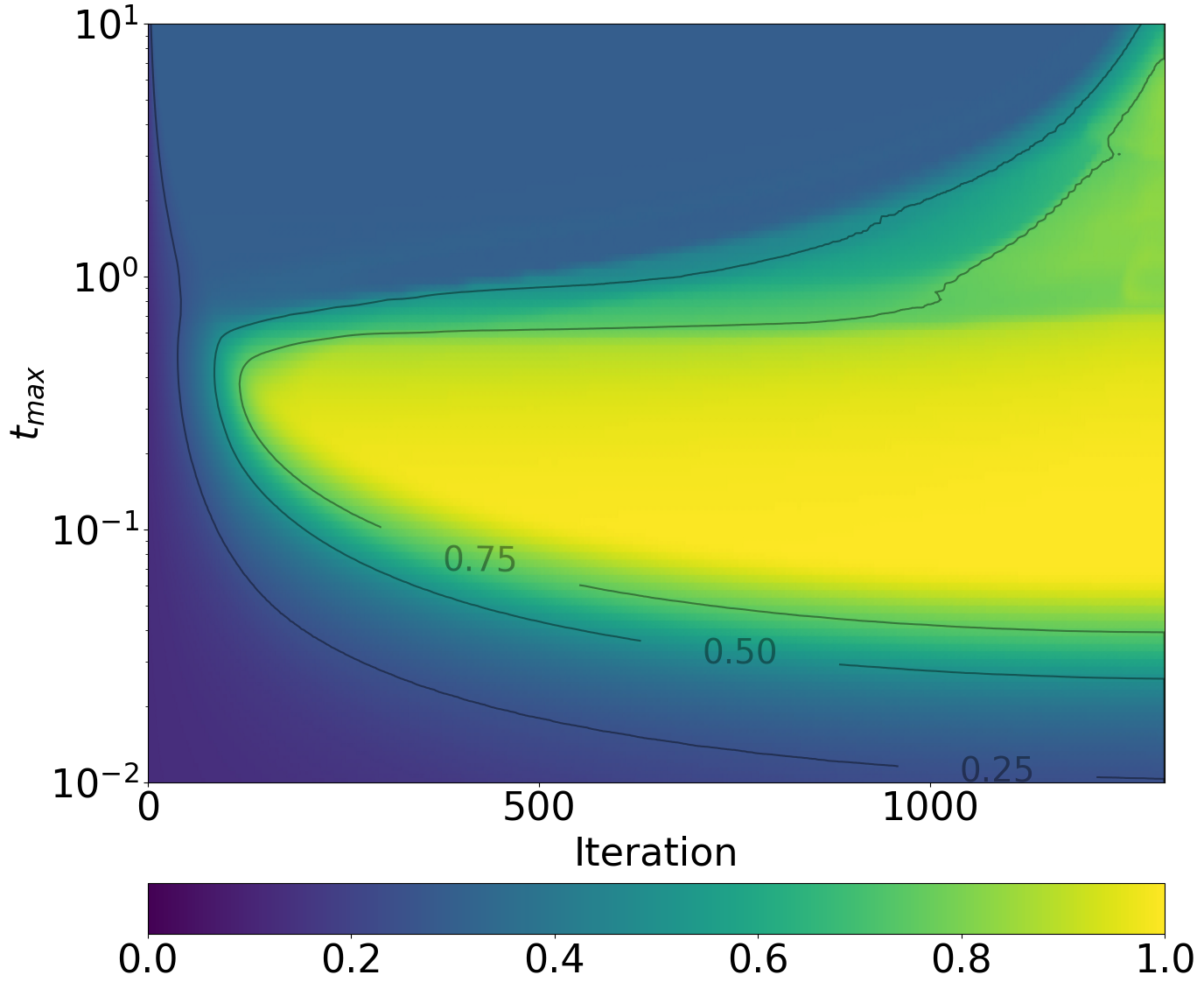}
        \caption{}
        \label{fig:2DIP:gradientLike:localMin}
    \end{subfigure}
    \hfill
    \caption{Convergence properties of the graduated non-convexity flow with constant step size (Algorithm \ref{alg:gradientLikeAlgorithm_v3}, Figures \ref{fig:2DIP:KoblerPock:xyplane}-\ref{fig:2DIP:KoblerPock:localMin}) and the gradient-like method with the adaptive smoothing schedule (Algorithm \ref{alg:gradientLikeAlgorithm}, Figures \ref{fig:2DIP:gradientLike:xyplane}-\ref{fig:2DIP:gradientLike:localMin}) for the 2D toy example. We show the rate of trajectories, for different choices of $\tmax$, converging to the global minimum depending on the initial starting point $\Bx_1$ (left column). Furthermore, we display the rate of trajectories which converge to the global minimum (middle column) and to stationary points (right column) depending on the initial smoothing parameter $\tmax$ and the iteration number.}
    \label{fig:2DIP}
\end{figure*}

\begin{theorem}\label{theorem:AdjustedGradientLike}
    Take $F$ and $f$, given in \eqref{eq:f} and \eqref{eq:seq_minimisation_1}, with $F(\cdot, t)$ continuously differentiable for $t\in [\tmin, \tmax]$ with $\tmin > 0$. Assume $\{t_i\}_{i\in \N}$ satisfy \eqref{eq:t_i}. Let $\{\Bd_i\}_{i\in \N},\ \{\Bx_i\}_{i\in \N}\subseteq \R^n$ be sequences generated with Algorithm \ref{alg:gradientLikeAlgorithm}, where $\Bd_i$ is determined using the adaptive smoothing schedule. 
    Moreover, we assume that $\alpha_\cdot \nabla_\Bx \mathcal{R}(\Bx, \cdot)$ is Lipschitz continuous with a global Lipschitz constant $L\geq 0$ for all $\Bx\in \R^n$.
    Then every limit point of the sequence $\{\Bx_i\}_{i\in \N}$ is a stationary point of $f.$
\end{theorem}

In case of the graduated non-convexity flow for the function \eqref{eq:seq_minimisation_1} and iterations \eqref{eq:graduated nonConvexity Flow},
the directions $\Bd_i$ correspond to
\begin{align}\label{eq:d_i}
    \Bd_i 
    &= - t_i \left( \bA^*(\bA\Bx_i - \By^\delta) + \alpha_{t_i} \nabla_\Bx \mathcal{R}(\Bx_i, t_i) \right),
\end{align}
which do not necessarily satisfy the descent condition 
\begin{align}
    \label{eq:gl_like}
    \langle \nabla_\Bx f(\Bx_i), \Bd_i\rangle_{\R^n}<0,
\end{align}
needed for convergence. We can address this issue through an adaptive smoothing schedule by selecting, in each iteration, the largest smoothing parameter $t_{j^\ast}$ such that the descent condition is satisfied. Such an index $j^\ast$ always exists as long as $\Bx_i$ is not a stationary point of $f$, since the maximal index $j^* = I$ in Algorithm \ref{alg:gradientLikeAlgorithm} leads to $t_{j^*} = \tmin$ and
\begin{equation*}
     \Big\langle \nabla_\Bx f(\Bx_i), - t_{j^*} \nabla_\Bx F(\Bx_i, t_{j^*})\Big\rangle_{\R^n} = -\tmin \Vert \nabla_\Bx f(\Bx_i) \Vert^2 <0.
\end{equation*}

Theorem \ref{theorem:AdjustedGradientLike} can be shown using this adaptive smoothing schedule and by ensuring the properties \ref{def:GradientLikeDirection:a} and \ref{def:GradientLikeDirection:b} in Definition \ref{def:GradientLikeDirection} hold. 

\subsection{Energy-based Parametrisation}
\label{sec:energy_based_param}
Adaptive step-size iterative methods require evaluating the objective function in each iteration to ensure convergence.
However, traditional SGMs approximate only the score function $\nabla_\Bx \log p_t(\Bx)$ and do not offer access to the likelihood. 
The alternative are energy-based models (EBM) \cite{hinton2002training}, where the probabilistic model is parametrised as $p_\theta(\Bx) = \frac{e^{g_\theta(\Bx)}}{Z(\theta)}$, for a scalar-valued neural network $g_\theta$ and a normalisation constant $Z(\theta)$. EBMs can be trained using the score matching objective by defining $s_\theta(\Bx_t, t) = \nabla_{\Bx_t} g_\theta(\Bx_t, t)$ \cite{salimans2021should}, which can be evaluated using common automatic differentiation libraries. This comes with a higher computational cost than the score parametrisation. 
Following Du et al. \cite{du2023reduce}, we parametrise the energy as \vspace{-1ex}
\begin{align}
    g_\theta(\Bx_t, t) = - \frac{1}{2} \| h_\theta(\Bx_t, t) \|_2^2, 
\end{align}\\[-3ex]
where $h_\theta$ is implemented as a time-conditional U-Net \cite{dhariwal2021diffusion}. The perturbed cost function is then given as \vspace{-1ex}
\begin{align}
    F(\Bx, t) = \frac{1}{2} \| \mathbf{A} \Bx - \By^\delta \|_2^2 + \frac{\alpha_t}{2} \| h_\theta(\Bx, t) \|_2^2.
\end{align}
\begin{remark}Likelihood in the EBM framework is defined by \vspace{-1ex}
\begin{align}
    \log p_\theta(\Bx_t, t) = g_\theta(\Bx_t, t) - \log Z(\theta, t).
\end{align}\\[-3ex]
This introduces a time step dependency into the (intractable) normalisation constant $Z(\theta, t)$. 
However, to compute the step sizes we only need to evaluate the objective up to additive constants, which means that $Z(\theta, t)$ does not need to be computed.
Energy-based parametrisation was used in a similar fashion to compute Metropolis correction probabilities \cite{du2023reduce}. 
\end{remark}

\section{Experiments}
In this section we investigate the numerical performance of Algorithm 2. We start with a toy example, where the data distribution is given by a Gaussian mixture model and the score can be computed analytically. For the second experiment we investigate computed tomography reconstruction on two datasets with a trained SGM. We use the PSNR and SSIM \cite{wang2004image} to evaluate the reconstruction quality.

\subsection{2D Toy Example}
\label{sec:toy_example}
To illustrate the behaviour and the convergence properties, we consider a Gaussian mixture model 
consisting of five Gaussians.
The density at time step $t$ is a Gaussian mixture model with a perturbed covariance matrix $\Sigma_k^t = \Sigma_k + \frac{\sigma^{2t} -1}{2 \log \sigma} \mathbf{I}$.
The diffusion process is given by the forward SDE $d\Bx_t = \sigma^t d \mathbf{w}_t$ with perturbation kernel $p_{t|0}(\Bx_t|\Bx_0)= \mathcal{N}(\Bx_t | \Bx_0, \frac{\sigma^{
2t} -1}{2 \log \sigma} \mathbf{I})$. 
Furthermore, we consider a simple two dimensional inverse problem with the forward operator $\mathbf{A} = \begin{pmatrix}
    1 & 1 \\ 
    0 & 0
\end{pmatrix}$ and clean measurements $\mathbf{y} = (2, 0)^\intercal.$ We choose a constant regularisation parameter $\alpha_t = 5$ and adjust one mean of the Gaussian mixture model to the position $\Bx^* = (1, 1)^\intercal$ in order to ensure that the global minimum of the cost function $f$ with $\tmin=10^{-3}$ is at $\Bx^*.$
We evaluate the graduated non-convexity flow (Algorithm \ref{alg:gradientLikeAlgorithm_v3}) with a constant step size $\lambda_i = 1$ as well as the gradient-like method (Algorithm \ref{alg:gradientLikeAlgorithm}) with the adaptive smoothing schedule. The values $t_i$ are evenly spaced between $\tmin$ and $\tmax.$ The goal is to analyse the algorithms in terms of their convergence properties with respect to the initialisations $\Bx_1,$ the initial smoothing parameter $\tmax$ and the iteration number. To do so, we run the algorithms with 1300 iterations for $10^{4}$ equally spaced initial points $\Bx_1$ on $[-10, 10]^2$ as well as 100 different values of $\tmax\in [10^{-2}, 10]$, which are evenly distributed on a log scale. The resulting rate of trajectories converging to stationary points and the global minimum are shown as a pseudocolor plot in Fig.~\ref{fig:2DIP}. The dependence on the initialisations is shown by the left column, where isolines display the loss landscape of $f$ and the orange paths of iterates show exemplary trajectories for different $\Bx_1$ and $\tmax.$ The dependence on the iteration number and $\tmax$ is shown by the middle and right column.




\subsection{Computed Tomography}
\begin{figure}[t]
    \centering
    \includegraphics[width=1.0\linewidth]{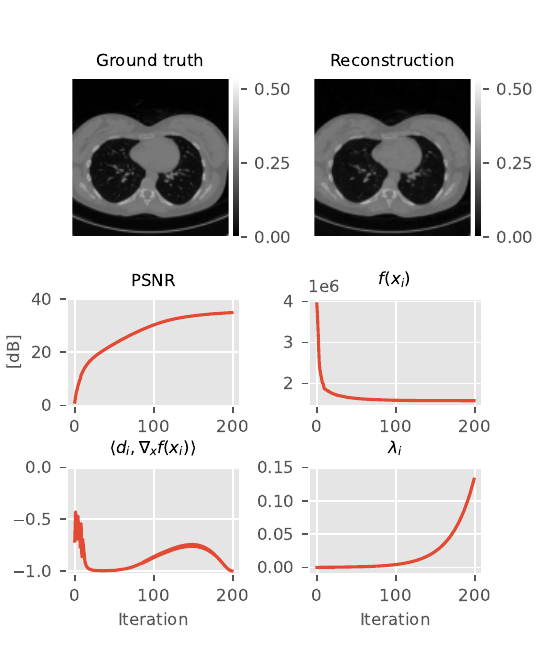}
    \caption{Result of the gradient like method for one example from the \textsc{AAPM} dataset. We show the PSNR, the objective value, the gradient-like condition and the computed step size during the iterations. The gradient-like condition is always satisfied during the iterations.}
    \label{fig:iteration}
\end{figure}

\begin{figure}[t]
     \centering
     \includegraphics[width=0.45\textwidth]{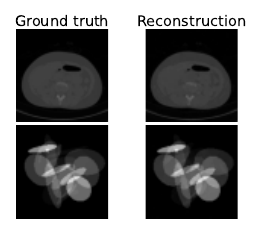}
     \caption{Example Reconstruction of \textsc{Ellipses} and \textsc{AAPM}. All images are shown in a color range $[0,1]$.}
     \label{fig:example_reco}
\end{figure}

We evaluate our approach on two datasets: \textsc{Ellipses} \cite{barbano2022educated} and \textsc{AAPM} \cite{mccollough2017low}. \textsc{Ellipses} contains $128\times128$px images of a varying number of ellipses with random orientation and size. The \textsc{AAPM} dataset consists of CT images from $10$ patients. We use images of $9$ patients to train the SGM, resulting in $2824$ training images. We use the remaining patient (id C027) for evaluation. 
We simulate measurements using a parallel-beam Radon transform with $60$ angles and use $10\%$ and $5\%$ relative Gaussian noise, respectively. For both datasets, we train SGMs using the VPSDE \cite{song2021scorebased}. We use a backtracking line search using the Armijo-Goldstein condition to determine a suitable step size with Barzilai-Borwein method~\cite{barzilai1988two} to find a candidate step size. We choose the parameter as $\alpha_t = \alpha \nu_t/\gamma_t$, where $\nu_t$ is the standard deviation and $\gamma_t$ the mean scaling of the perturbation kernel. The value $\alpha$ was set according to a coarse sweep over one example image. In Fig.~\ref{fig:iteration} we show the result for the gradient like method on one \textsc{AAPM} example, including the PSNR, objective value, step size and gradient-like condition from \eqref{eq:gl_like}. Fig.~\ref{fig:example_reco} shows example reconstructions for both datasets.

\paragraph*{Initial value}
The gradient-like method (Alg.~\ref{alg:gradientLikeAlgorithm_v3}) is deterministic, except for the choice of initialisation $\Bx_1$. We study the effect of $\Bx_1$ for both \textsc{Ellipses} and \textsc{AAPM}, see Table~\ref{tab:ellipses_aapm_initial_choice} by comparing Alg.~\ref{alg:gradientLikeAlgorithm_v3} with gradient descent for optimising the objective function at $\tmin$, see \eqref{eq:f}. For both datasets we use $300$ steps with a logarithmic time interval for Alg.~\ref{alg:gradientLikeAlgorithm_v3}. We evaluate $10$ random initialisations $\Bx_1$ for $10$ images. 
We compute mean PSNR and SSIM values over all reconstructions. Further, we compute the mean standard deviation value over images. We see that for both \textsc{Ellipses} and \textsc{AAPM} the mean PSNR and mean SSIM are higher, while the mean standard deviation is lower. That is, the gradient-like method achieves better reconstructions more consistently.\vspace{-2ex}

\begin{table}[t]
\centering
\caption{Mean and mean standard deviation of PSNR and SSIM for $10$ reconstructions on the \textsc{Ellipses} and \textsc{AAPM} dataset for $10$ different starting values.}
\begin{tabular}{llccc}
\toprule
 &   & & Gradient Descent  &  Alg.~\ref{alg:gradientLikeAlgorithm_v3} \\ \midrule
\multirow{4}{*}{\rotatebox[origin=c]{90}{\parbox[c]{1.4cm}{\centering \textsc{Ellipses}}}}  & \multirow{ 2}{*}{\textbf{PSNR}} & mean & $28.96$   & $32.93$  \\
& & std & $0.43$ & $0.10$ \\[0.1cm] 
& \multirow{ 2}{*}{\textbf{SSIM}} & mean &  $0.771$  & $0.933$  \\
& & std & $0.019$ & $0.001$ \\ \midrule
\multirow{4}{*}{\rotatebox[origin=c]{90}{\parbox[c]{1.2cm}{\centering \textsc{AAPM}}}} & \multirow{ 2}{*}{\textbf{PSNR}} & mean &  $35.89$  & $37.33$  \\
& & std & $0.032$ & $0.028$ \\[0.1cm] 
& \multirow{ 2}{*}{\textbf{SSIM}} & mean &  $0.899$  & $0.913$  \\
& & std & $0.032$ & $0.0004$\\\bottomrule 
 
\end{tabular}
\label{tab:ellipses_aapm_initial_choice}
\end{table}


\begin{table}[]
\centering
\caption{PSNR and SSIM for $100$ reconstructions on the \textsc{AAPM} dataset for a constant step size ($1200$ iterations) and adaptive step size ($300$ iterations).}
\begin{tabular}{lcccc}
\toprule
     & \multicolumn{2}{c}{AAPM} & \multicolumn{2}{c}{Ellipses} \\
     & \textbf{PSNR} & \textbf{SSIM}  & \textbf{PSNR}          & \textbf{SSIM}         \\ \midrule
Constant step size & $32.93$  & $0.748$  & $30.82$  & $0.742$ \\
Adaptive step size &  $37.08$  & $0.896$ &  $33.21$ &  $0.935$  \\ \bottomrule 
\end{tabular}
\label{tab:aapm_step_size}
\end{table}

\paragraph*{Adaptive step size} The energy-based parametrisation comes with a higher computational cost. However, it allows uing adaptive step sizes for Alg.~\ref{alg:gradientLikeAlgorithm}. In Table~\ref{tab:aapm_step_size} we show that this leads to a better reconstruction than a fixed step size. Compare also the results in Fig.~\ref{fig:iteration}, which shows that the computed step size changes by several orders of magnitude over the iterations. Also, we need fewer iteration ($300$ vs $1200$) to get convincing reconstructions.

\section{Discussion}
In the toy example in Section \ref{sec:toy_example} it seems that the gradient-like condition in Definition~\ref{def:GradientLikeDirection} hinders the ability of the algorithm to reach the global optimum, as the search direction $\Bd_i$ always points in the same halfspace as $-\nabla_\Bx f(\Bx_i)$. One possible approach to circumvent this is to only enforce this condition after a set number of iterations in order to ensure the convergence to a stationary point of $f$. 
However, for the high-dimensional imaging experiments the gradient-like condition does not seem to be too restrictive. As we see in Fig.~\ref{fig:iteration}, the condition is satisfied for all iterations and we still achieve a good reconstruction. Further, the adaptive step size rule leads to a improvement in terms of both PSNR and SSIM, while using fewer iterations, which warrants the additional computational cost of the step size search.

\section{Conclusion}
In this work we show that SGMs can be used in a graduated optimisation framework to solve inverse problems. Further, we propose to use an energy-based parametrisation, which enables the use of line search method for finding a suitable step size. Using the theory of gradient-like directions, we can prove convergence to stationary points. We hypothesise that this method is also able to better escape saddle points and converge to local minima. This research question is left for further work.


\bibliographystyle{IEEEbib}
\bibliography{main}

\clearpage
\section*{Appendix}
\subsection*{Proof of Theorem~\ref{theorem:AdjustedGradientLike}}

\begin{manualtheorem}{3.2}
    Let $F$ and $f$ be given as in \eqref{eq:f} and \eqref{eq:seq_minimisation_1} and let $F(\cdot, t)$ be continuously differentiable for $t\in [\tmin, \tmax]$ with $ \tmax > \tmin > 0$. Furthermore, let $\{t_i\}_{i\in \N}$ be as in \eqref{eq:t_i} and $\{\Bd_i\}_{i\in \N}\subseteq \R^n$ as well as $\{\Bx_i\}_{i\in \N}\subseteq \R^n$ be the sequences generated based on the update rules of Algorithm \ref{alg:gradientLikeAlgorithm}, where $\Bd_i$ is determined by using the adaptive smoothing schedule. 
    Moreover, we assume that $\alpha_\cdot \nabla_\Bx \mathcal{R}(\Bx, \cdot)$ is Lipschitz continuous with a global Lipschitz constant $L\geq 0$ for all $\Bx\in \R^n$, i.e. there exists a constant $L\geq 0$ such that
    \begin{equation*}
        \Vert \alpha_t \nabla_\Bx \mathcal{R}(\Bx, t) - \alpha_{\tilde{t}} \nabla_\Bx \mathcal{R}(\Bx, \tilde{t})\Vert \leq L \vert t - \tilde{t} \vert
    \end{equation*}
    $\forall t,\tilde{t}\in [\tmin,\tmax],\ \forall \Bx\in \R^n.$
    Then every limit point of the sequence $\{\Bx_i\}_{i\in \N}$ is a stationary point of $f.$
\end{manualtheorem}

\begin{proof}
    As described in Section \ref{sec:Gradient-like Methods}, the adaptive smoothing schedule selects in the $i$-th iteration of Algorithm \ref{alg:gradientLikeAlgorithm} the largest smoothing parameter $t_{j^*}$ such that the descent condition in equation \eqref{eq:gl_like} with $\Bd_i = - \tilde{t}_i \nabla_\Bx F(\Bx_i, \tilde{t}_i)$ and $\tilde{t}_i \coloneqq t_{j^*}$ holds. In other words, we choose $j^*$ to be
    \begin{align*}
        j^*\coloneqq \min\Big\{j\in \mathbb{N} \ \Big| \ \Big\langle \nabla_\Bx f(\Bx_i), - t_j \nabla_\Bx F(\Bx_i, &t_j)\Big\rangle_{\R^n} < 0 \\ &\land\ t_j \leq \tilde{t}_{i-1} \Big\}.
    \end{align*}
    Such an index always exists as long as $\Bx_i$ is not a stationary point of $f$ (which is ensured by Algorithm \ref{alg:gradientLikeAlgorithm} due to the stopping criterion in Line 2), as we have
    \begin{align*}
        \lim_{j\to \infty} \Big\langle \nabla_\Bx f(\Bx_i), - t_j \nabla_\Bx F(\Bx_i, t_j)\Big\rangle_{\R^n} &= -\tmin \Vert \nabla_\Bx f(\Bx_i)\Vert^2 \\
        &<0,
    \end{align*}
    where the equation follows from the continuity of $\nabla_\Bx F(\Bx_i, \cdot).$ Hence, there exists a $N\in \mathbb{N},$ such that $$\Big\langle \nabla_\Bx f(\Bx_i), - t_j \nabla_\Bx F(\Bx_i, t_j)\Big\rangle_{\R^n} < 0$$ for all $j\geq N.$ The above choice of the $\Bd_i$ ensures that $\langle \nabla_\Bx f(\Bx_i), \Bd_i\rangle_{\R^n} < 0$ for all $i\in \N,$ which is a needed criterion for the gradient-like algorithm to converge according to \cite{geiger2013numerische}. The sequence $\{\tilde{t}_i\}_{i\in \N}$ is a subsequence of $\{t_i\}_{i\in\N}$ and fulfills the properties $\tilde{t}_{i+1} \leq \tilde{t}_i$ as well as $\lim_{i\to\infty} \tilde{t}_i = \tmin.$

    Proof by contradiction: Let $\Bx^*\in \R^n$ be a non-stationary point of $f$ and let $\{\Bx_{i_j}\}_{j \in \N}$ be a subsequence, which converges to $\Bx^*.$ The aim is to show both properties \ref{def:GradientLikeDirection:a} and \ref{def:GradientLikeDirection:b} in Definition \ref{def:GradientLikeDirection}. We would then obtain that $\{\Bd_i\}_{i\in \N}$ is gradient-like with respect to $f$ and $\{\Bx_i\}_{i\in \N}.$ Hence, by \cite[Theorem 8.9]{geiger2013numerische}, we would obtain that every accumulation point of $\{\Bx_i\}_{i\in \N}$ is a stationary point of $f,$ which would give the desired contradiction to the assumption that $\nabla_\Bx f(\Bx^*) \neq 0.$
    
    We first show that $\Vert \Bd_{i_j} \Vert\leq \varepsilon_1 \ \forall j\in \N,$ which corresponds to property \ref{def:GradientLikeDirection:a} of Definition \ref{def:GradientLikeDirection}. It holds that
    \begin{align*}
        &\Vert \Bd_{i_j} \Vert \\
        &= \Vert - \tilde{t}_{i_j} \nabla_\Bx F(\Bx_{i_j}, \tilde{t}_{i_j}) \Vert\\
        &= \Big\Vert \tilde{t}_{i_j} \left( \bA^*(\bA\Bx_{i_j} - \By^\delta) + \alpha_{\tilde{t}_{i_j}} \nabla_\Bx \mathcal{R}(\Bx_{i_j}, \tilde{t}_{i_j}) \right)\Big\Vert,\\
        &\leq \tmax \Big( \Vert \nabla_{\Bx}f(\Bx_{i_j}) \Vert \\
        &\hspace{8ex}+ \Vert \alpha_{\tilde{t}_{i_j}} \nabla_\Bx \mathcal{R}(\Bx_{i_j}, \tilde{t}_{i_j}) - \alpha_{\tmin} \nabla_\Bx \mathcal{R}(\Bx_{i_j}, \tmin) \Vert\Big)\\
        &\leq \tmax \left( \Vert \nabla_{\Bx}f(\Bx_{i_j}) \Vert + L \vert \tilde{t}_{i_j} - \tmin \vert \right)\\
        &\leq \tmax \left( \Vert \nabla_{\Bx}f(\Bx_{i_j}) \Vert + L (\tmax - \tmin) \right).
    \end{align*}
    As $f$ is continuously differentiable and with $\lim_{j\to \infty} \Bx_{i_j} = \Bx^*,$ we have that $\nabla_{\Bx}f(\Bx_{i_j}) \to \nabla_{\Bx}f(\Bx^*)$ for $j\to \infty$ and the sequence $\{\nabla_{\Bx}f(\Bx_{i_j})\}_{j\in \N}$ is bounded, i.e.\ there exists a constant $\tilde{c}\geq 0$ such that $\Vert \nabla_{\Bx}f(\Bx_{i_j}) \Vert \leq \tilde{c} \ \ \forall j\in \N$.
    Therefore, we have
    \begin{align*}
        \Vert \Bd_{i_j} \Vert &\leq \tmax \left( \tilde{c} + L (\tmax - \tmin) \right),
    \end{align*}
    which shows property \ref{def:GradientLikeDirection:a} of Defintion \ref{def:GradientLikeDirection}. To prove property \ref{def:GradientLikeDirection:b}, we compute
    \begin{align*}
        &\langle \nabla_\Bx f(\Bx_{i_j}), \Bd_{i_j}\rangle_{\R^n}\\
        &= -\tilde{t}_{i_j} \langle \nabla_\Bx f(\Bx_{i_j}), \nabla_\Bx f(\Bx_{i_j}) + \alpha_{\tilde{t}_{i_j}} \nabla_\Bx \mathcal{R}(\Bx_{i_j}, \tilde{t}_{i_j})  \\
        &\hspace{27ex} - \alpha_{\tmin} \nabla_\Bx \mathcal{R}(\Bx_{i_j}, \tmin) \rangle_{\R^n}\\
        &\leq -\tilde{t}_{i_j} \Vert \nabla_\Bx f(\Bx_{i_j}) \Vert^2 + \tilde{t}_{i_j} \Vert \nabla_\Bx f(\Bx_{i_j}) \Vert \cdot L \vert \tilde{t}_{i_j} - \tmin \vert\\
        &\leq -\tmin \Vert \nabla_\Bx f(\Bx_{i_j}) \Vert^2 + \tmax \tilde{c} L \vert \tilde{t}_{i_j} - \tmin \vert \eqqcolon s_{i_j}.
    \end{align*}
    As $\nabla_\Bx f$ is continuous and $\Bx_{i_j}\to \Bx^*$ for $j\to \infty$ with $\Bx^*$ being a non-stationary point of $f,$ we have that
    \begin{equation*}
        \lim_{j\to\infty} \Vert \nabla_\Bx f(\Bx_{i_j}) \Vert^2 = \Vert \nabla_\Bx f(\Bx^*) \Vert^2 >0.
    \end{equation*}
    The sequence $s_{i_j}$ converges and it follows
    \begin{equation*}
        \lim_{j\to\infty} s_{i_j} = -\tmin  \Vert \nabla_\Bx f(\Bx^*) \Vert^2 < 0.
    \end{equation*}
    By definition of the convergence, it holds $\forall \varepsilon\!>\! 0 \ \exists N(\varepsilon)\geq 0:$
    \begin{equation*}
        \vert s_{i_j} - (-\tmin \Vert \nabla_\Bx f(\Bx^*) \Vert^2) \vert < \varepsilon \quad \forall j\geq N(\varepsilon).
    \end{equation*}
    By choosing $\varepsilon \coloneqq \tmin \Vert \nabla_\Bx f(\Bx^*) \Vert^2/2>0,$ we obtain in particular
    \begin{equation*}
        s_{i_j} < - \tmin \Vert \nabla_\Bx f(\Bx^*) \Vert^2/2 < 0 \quad \forall j\geq N(\varepsilon),
    \end{equation*}
    which shows property \ref{def:GradientLikeDirection:b} of Definition \ref{def:GradientLikeDirection}.
\end{proof}

\end{document}